\documentclass[letterpaper, 10 pt, journal, twoside]{IEEEtran}
\usepackage{graphicx}
\usepackage{setspace}
\usepackage{amsmath, nccmath}
\usepackage{amssymb}
\usepackage{mathtools}
\DeclareMathOperator*{\argmin}{arg\,min}
\usepackage{url}
\usepackage{soul}
\usepackage{color}
\usepackage{subcaption}
\usepackage{mwe}
\usepackage{microtype}
\DeclareMathOperator*{\argmax}{arg\,max}
\usepackage{hyperref}
\usepackage{bbold}
\ifCLASSINFOpdf
\else
\fi
\begin{document}
%
\title{A Framework for Learning from Demonstration with Minimal Human Effort}
%
%
%

\author{Marc Rigter$^{1}$, Bruno Lacerda$^{1}$, and Nick Hawes$^{1}$%
\thanks{Manuscript received: September, 10, 2019; Revised December, 18, 2019; Accepted January, 22, 2020.}
\thanks{This paper was recommended for publication by Editor Dongheui Lee upon evaluation of the Associate Editor and Reviewers' comments.} 
\thanks{$^{1}$Marc Rigter, Bruno Lacerda, and Nick Hawes are with the Oxford Robotics Institute, University of Oxford, United Kingdom ({email: \tt\small \{mrigter, bruno, nickh\} @robots.ox.ac.uk})}
\thanks{Digital Object Identifier (DOI): see top of this page.}
}
%
%

\markboth{IEEE Robotics and Automation Letters. Preprint Version. Accepted January, 2020}
{Rigter \MakeLowercase{\textit{et al.}}: A Framework for Learning from Demonstration with Minimal Human Effort} 

%



\maketitle

\begin{abstract}
We consider robot learning in the context of shared autonomy, where control of the system can switch between a human teleoperator and autonomous control. In this setting we address reinforcement learning, and learning from demonstration, where there is a cost associated with human time. This cost represents the human time required to teleoperate the robot, or recover the robot from failures. For each episode, the agent must choose between requesting human teleoperation, or using one of its autonomous controllers. In our approach, we learn to predict the success probability for each controller, given the initial state of an episode. This is used in a contextual multi-armed bandit algorithm to choose the controller for the episode. A controller is learnt online from demonstrations and reinforcement learning so that autonomous performance improves, and the system becomes less reliant on the teleoperator with more experience. We show that our approach to controller selection reduces the human cost to perform two simulated tasks and a single real-world task.
\end{abstract}


%
\IEEEpeerreviewmaketitle

\section{Introduction}
%
%
%
%
\IEEEPARstart{I}{ntegrating} demonstrations with Reinforcement Learning (RL) has been applied to a number of difficult problems in sequential decision making and control. Examples include Atari games~\cite{hester2018deep} and manipulation tasks~\cite{rajeswaran2018learning}~\cite{nair2018overcoming}. In most settings, it is assumed that there is a fixed set of demonstration data. In this work, we are interested in \textit{shared autonomy}, where control may switch back and forth between a human teleoperator, and autonomous control. This is common in many domains in which fully autonomous systems are not yet viable, and human intervention is required to increase system capability, or recover from failures (e.g.~\cite{dorais1998adjustable}~\cite{hawes2017strands}~\cite{chiou2016experimental}). In shared autonomy it is desirable to reduce the burden on the human operator by only handing control to the human in situations where autonomous performance is poor, and only handing control to the robot when autonomous performance is good. The autonomous experience and human demonstrations may be used to improve autonomous performance so that the system becomes less reliant on the human for future tasks.

Elements of this problem have been addressed before. Optimally switching control in shared autonomy has been approached using Markov Decision Process (MDP) planning~\cite{wray2016hierarchical}~\cite{jansen2017synthesis}. However, most planning approaches assume a known model for the performance of the human and autonomous system, which may not be available in reality, or may change over time. Requesting human input has been approached with the \textit{ask for help} framework, where the agent asks for a human demonstration if uncertainty is above a hand-tuned threshold~\cite{clouse1996integrating}~\cite{del2018not}~\cite{chernova2009interactive}~\cite{lin2017explore}. Unlike the ask for help approaches, we seek to explicitly minimise a cost function associated with human exertion, and unlike the planning approaches we do not assume models are known a priori, and learn a control policy online. 

\begin{figure}[t!]
	\centering
	\includegraphics[trim={0cm 0cm 0cm 0cm},clip,width=0.38\textwidth]{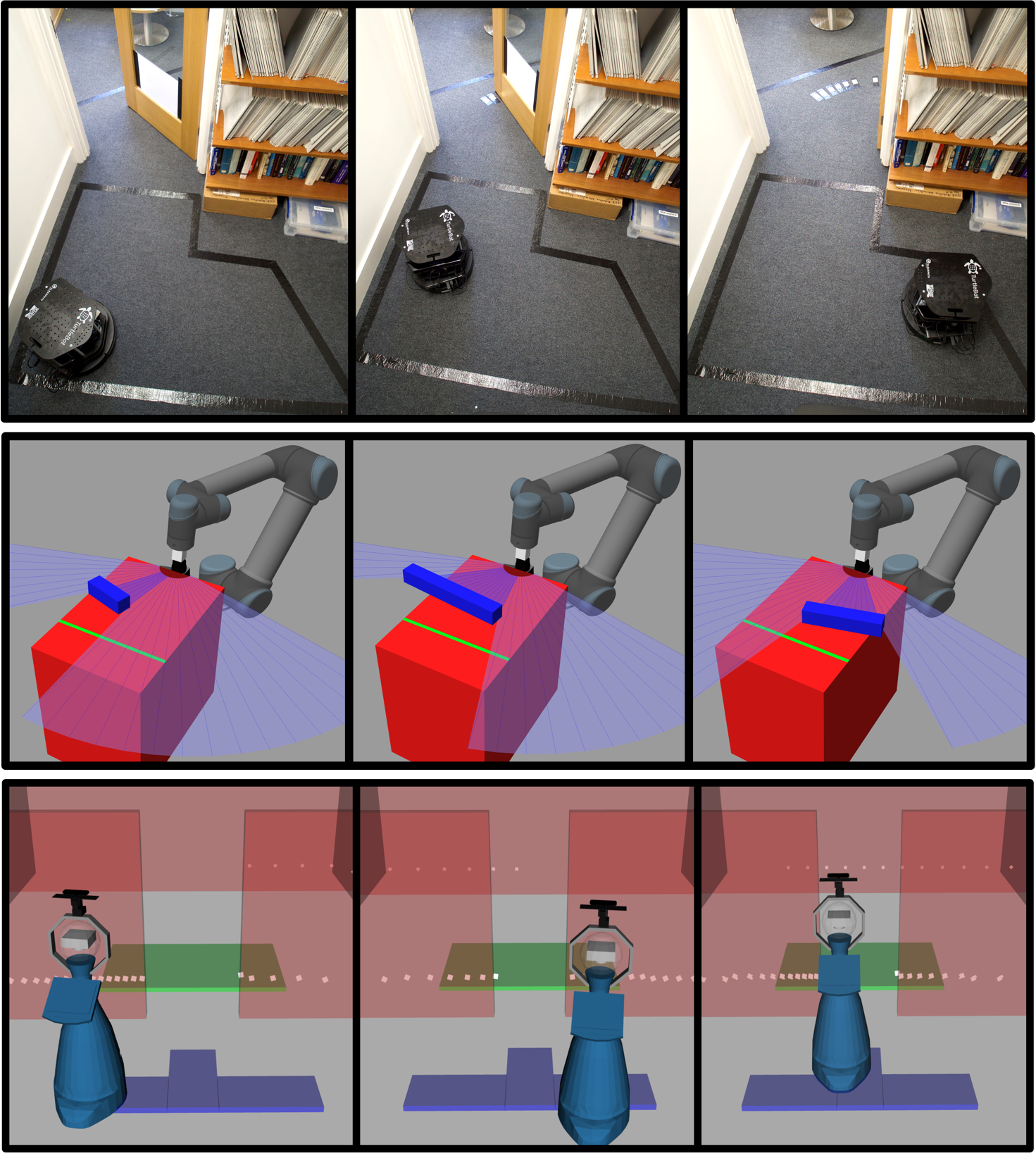} 
	\caption{Reprensentative initial configurations in our evaluation domains: real-world navigation (top), block pushing (middle) and simulated navigation (bottom).\label{fig:domains}}
	\vspace{-5mm}
\end{figure}

This work considers episodic problems with a binary outcome of success or failure. We assume that there is a cost for asking the human for a demonstration, and a cost for failing an episode. These costs represent the time required for the human to perform a demonstration, or to recover the robot from failure. Motivating examples include robot navigation, where a human assists the robot when it is stuck~\cite{hawes2017strands}, and manipulation, where a human recovers items dropped by the robot~\cite{akkaladevi2016towards}. For each episode, the system must decide whether to ask for a human demonstration, or give control to one of the available autonomous controllers. We assume that an autonomous controller is learnt online as the system gathers more experience. We also allow for additional autonomous controllers, as in many domains there exist pre-programmed controllers which can be utilised to reduce operator workload. 

The contributions of this paper are a general framework for this problem, and a specific instantiation of the framework. In the framework, controller choice is formulated as a contextual Multi-Armed Bandit (MAB). In our instantiation, we use continuous correlated beta processes to estimate the MAB probabilities. A control policy is learnt online from demonstrations plus RL so that performance improves and the system becomes increasingly autonomous. We evaluate our approach on three domains (Figure~\ref{fig:domains}),  including experiments on a real robot, and show empirically that our approach reduces the total human cost relative to other approaches.

%

\section{Related Work}
Previous research has combined human input with RL to guide the learning process. In supervised actor-critic RL~\cite{barto2004j}, the action applied at each time step is a weighted sum of the actions suggested by a human supervisor and by the learner. In interactive RL, the reward signal is given by a human \cite{suay2011effect}. 

Other approaches modify the control input of the human to improve performance whilst maintaining the control authority of the operator. Reddy et al.~\cite{reddy2018deep} apply deep RL to learn an approximate Q-function. The action applied to the system is the action near the teleoperator input with highest value.~\cite{schwarting2017parallel} uses Model Predictive Control (MPC) to determine the optimal control from a known model, with a cost for deviating from the user control inputs.~\cite{Broad2017LearningMF} also uses MPC, but learns the model from data collected from human-machine interaction. All of the above approaches require that the human operator is present at all times, while we are interested in methods which reduce the input required from the operator.

The ask for help framework introduced in~\cite{clouse1996integrating} only asks for human input when the learner is uncertain. In~\cite{clouse1996integrating}, the agent nominally performs tabular Q-learning. If all actions in a state have a similar Q-value, the agent is deemed uncertain, and asks the human which action to take. A similar idea is applied in the Deep Q-Network (DQN) setting in~\cite{lin2017explore}, where the loss of the DQN is used to gauge uncertainty. Chernova and Veloso~\cite{chernova2009interactive} consider policies defined by a support-vector machine classifier. For states near a decision boundary of the classifier, a demonstration is requested. In \cite{del2018not}, Gaussian process policies are learnt from demonstrations. A demonstration is requested if the variance of the action output is too high. In contrast to our work, all of these approaches require a hand-tuned threshold for uncertainty.

In \textit{predictive advising}~\cite{torrey2013teaching}, the teacher learns to predict the action that the learner will take in a given state. In a new state, if the predicted action does not equal the ``correct'' action known by the teacher, the teacher chooses the action taken. This method would be difficult to apply to continuous action spaces where it is unclear whether an action is correct.

MDP planning has been applied to the problem of optimally switching control between a human and agent. In~\cite{fu2016synthesis}, the authors find Pareto-optimal MDP policies which trade off minimising a cost function for operator effort, and maximizing the probability of success of satisfying a temporal logic specification. Such a cost representing human inconvenience is often referred to as ``bother'' cost~\cite{cohen2011user}. Wray et al.~\cite{wray2016hierarchical} consider that the human and autonomous system have different known capabilities, and generate plans such that the system never enters states where the entity in control cannot act. These approaches assume a model is known for the human and autonomous agent. Lacerda et al.~\cite{lacerda2019probabilistic} plan over an MDP with transition probabilities learnt from executing each transition many times~\cite{fentanes2015now}. This approach does not require prior knowledge, but is limited to long-term deployment in the same environment. The authors use human intervention to recover from failures~\cite{hawes2017strands}, but do not plan for human control.


Probabilistic policy reuse~\cite{fernandez2006probabilistic} assumes access to a library of baseline policies which may be reused, overriding the actions of the current policy with some probability. The algorithm maintains an average of the return from reusing each policy. Higher returns increase the likelihood a policy is chosen. This approach determines which baseline policy is most useful for learning the task. In our work, we consider variable controller performance throughout the state space, and choose the controller depending on the initial state.

A comparison can be made between our work and hierarchical RL~\cite{sutton1999between} which uses \emph{options}. An option is a temporally extended action, similar to the choice of controller in our work. In hierarchical RL, the policy which chooses options is also learnt with RL algorithms. In contrast, we consider controller choice as a contextual MAB and use uncertainty estimates to guide exploration towards promising choices. 

In contrast to previous work, we both explicitly minimise a cost function associated with the human workload, and do not assume prior knowledge of models of performance.

\section{Preliminaries}

\subsection{Multi-Armed Bandits}
In a Multi-Armed Bandit (MAB), at each episode an agent must choose from a finite set $\Phi$ of arms, with unknown reward distributions $r_\phi$ for choosing each arm $\phi \in \Phi$. The objective is to maximise the cumulative reward received. Algorithms which solve the MAB problem must balance exploration to gather information about the arms with exploiting the best known arm.
In this work, we consider an extension of the MAB that is \emph{non-stationary} and \emph{contextual.} 

%
%

In a contextual MAB~\cite{li2010contextual}, at episode $k$ the agent also receives information about the ``context'' in the form of a state $s_{k}$, which may inform the agent about the reward distribution of each arm for episode $k$. Over many trials, algorithms for the contextual MAB estimate the mean and variance of the reward for arm $\phi \in \Phi$, given state $s_k$ where ${\mu(\phi, s_{k}) = \mathbb{E}[r_\phi | s_k]}$, and ${\sigma(\phi, s_k)^2 = var[r_\phi | s_k]}$. One effective approach to arm selection is to use an Upper Confidence Bound (UCB) algorithm~\cite{li2010contextual}:

\begin{equation}
\label{eqn:init_bandit}
\phi_k = \argmax_{\phi \in \Phi} \Big(\mu(\phi, s_k) + \alpha \sigma(\phi, s_k) \Big),
\end{equation}

\noindent where $\alpha$ trades off exploration versus exploitation.

Our contextual MAB may also be non-stationary with the reward distribution changing between episodes (eg. due to learning). A simple approach to address this is to use a \emph{sliding window} and only use the $m$  most recent trials to estimate the distributions in Equation~\ref{eqn:init_bandit}~\cite{garivier2008on}.

\subsection{Continuous Correlated Beta Processes}
We wish to choose a function approximator to predict the probabilities required in the MAB. To estimate probabilities in the range [0,1], the output of a Gaussian Process (GP) can be passed through a logistic function to form a Logistic GP (LGP). However, for LGPs, the posterior can only be computed approximately, and computation is cubic with the size of the dataset. A simpler alternative is the Continuous Correlated Beta Process (CCBP), proposed in~\cite{goetschalckx2011continuous}. Like LGPs, CCBPs are a nonparametric function approximator with a model of uncertainty, and a range of [0,1], but have computation time linear with the number of data points. 

Let $S$ be a continuous state space and ${\mathcal{B} = \{B_s \ |\ s \in S \}}$ the space of Bernoulli trials indexed by states of $S$, with unknown success probability ${p(s) = \Pr(B_s = 1)}$. The outcome, $o \in \{0,1\}$ of each trial may either be successful, denoted 1, or unsuccessful, denoted 0. The distribution over $p(s)$ after $\alpha(s)$ outcomes of success and $\beta(s)$ outcomes of failure is given by a beta distribution:

\begin{equation}
\Pr(p(s)) = Beta(\alpha(s), \beta(s)) \propto p(s)^{\alpha(s) - 1}(1-p(s))^{\beta(s)-1}
\end{equation}

To obtain accurate values for $p(s)$ over the entire state space, we would need to observe many outcomes at each $s$, which is not possible in continuous space. In a CCBP we assume $p(s)$ is a smooth function, such that the Bernoulli trials are correlated, and experience can be shared between them. A kernel function, $K(s, s') \in [0,1]$ is used to determine the extent to which experience from trial $B_s$ should be shared with $B_{s'}$.
Consider the set  ${O=  \{B_{s_0} = o_0, B_{s_1} = o_1,\ldots, B_{s_{T}} = o_{T}\}}$ of observed outcomes $o_0, \ldots, o_T$ after running $T+1$ different trials.
The posterior beta distribution for an experiment $B_s$ is given by:

\begin{multline}
\Pr(p(s)|O) \propto p(s)^{\alpha(s) - 1 + \sum_{t=0}^{T} o_t K(s_t, s)} \\
\times  (1-p(s))^{\beta(s) - 1 + \sum_{t=0}^{T} (1 - o_t) K(s_t, s)}.
\label{eqn:ccbp}
\end{multline}

\subsection{Deep Deterministic Policy Gradients}
In an RL problem, an agent must learn to act by receiving a reward signal corresponding to performing an action $a \in A$ at a state $s \in S$. At discrete time step $t$, the agent takes action $a_t$ according to a policy, $\pi:S\rightarrow A$. We define the return $G_t = \sum_{i=t}^T \gamma^{(i-t)} r_i$ where $T$ is the horizon, $\gamma < 1$ is the discount factor, and $r_i$ is the reward received at time step $i$. We consider episodic problems in which $T$ is the end of the episode. The goal of RL is to find $\pi$ which maximises the expected return, $J = \mathbb{E}_\pi [G_t]$.

Deep Deterministic Policy Gradients (DDPG)~\cite{lillicrap2015continuous} is an off-policy, model-free, RL algorithm capable of utilizing neural network function approximators. DDPG maintains a state-action value function, or critic, $Q(s, a)$, with parameters $\theta^Q$, and a policy, or actor, $\pi(s)$, with parameters $\theta^\pi$. Additionally, a replay buffer $R$ of recent transitions experienced is maintained containing tuples of $(s_t, a_t, r_t, s_{t+1})$. 

The algorithm alternates between collecting experience by acting in the environment, and updating the actor and critic. For exploration, noise is added to the actions chosen by the actor during training: $a_t = \pi(s_t) + \mathcal{N}$, where $\mathcal{N}$ is a noise process. During each update step, a minibatch of $N$ samples is taken from $R$ to update the actor and critic functions. The critic parameters, $\theta_Q$, are updated to minimise the loss

\begin{equation}
L = \frac{1}{N}\sum_i(y_i - Q(s_i, a_i | \theta_Q))^2,
\end{equation}

\noindent where the targets $y_i$ are computed from the Bellman equation: 
\begin{equation}
y_i = r_i + \gamma Q(s_{i+1}, \pi(s_{i+1}|\theta^\pi)|\theta^Q).
\label{eqn:target}
\end{equation}

The actor is updated using the deterministic policy gradient:
\begin{equation}
\nabla_{\theta^\pi}J = \frac{1}{N} \sum_i \nabla_a Q(s, a|\theta^Q)|_{s=s_i, a=\pi{(s_i)}}\nabla_{\theta^\pi}\pi(s|\theta^\pi)|_{s=s_i}
\end{equation}
Intuitively, this update changes the actor parameters to produce actions with a higher $Q$-value as judged by the critic. The target values in Equation \ref{eqn:target} are usually computed using separate networks for the actor and critic whose weights, $\theta^{\pi'}$ and $\theta^{Q'}$, are an exponential average over time of $\theta^\pi$ and $\theta^Q$ respectively. This is necessary to stabilise learning.

\section{Methodology}
We assume that our system must perform an episodic task. Each episode $k$ has an initial state, $s_{k,0} \in S_0$, and a binary outcome, $o_k \in \{0,1\}$, of success for reaching a goal state $s_g \in S_g$, or failure. For each episode a controller must be chosen. The controller may be human teleoperation, which we denote $C_h$, or one of the $n$ available autonomous controllers, $C_{a,i}$. The autonomous controllers may be pre-programmed, or learnt online. We assume that there is a cost, $c_d$, per human demonstration representing the human time to give a demonstration, and a cost $c_f$ for failure, representing the human time required to recover the robot from failure. 
For the $k^{th}$ episode, the system chooses a controller, ${C_k \in \mathcal{C}}$, where $\mathcal{C} = \{C_h, C_{a,i}, \ldots, C_{a,n}\}$. The objective is to minimise the cumulative cost $\sum c_h$, defined as the sum of the demonstration and failure costs over the episodes. 


In the next subsection, we describe our general approach for controller selection to minimise the cumulative cost.  Controller selection is considered as a contextual MAB, and an upper-confidence bound algorithm is applied to choose the controller at each episode. We then describe how the CCBP can be used for the performance prediction aspect of our framework, and how a controller can be learnt using DDPG with demonstrations so that the autonomous performance improves, making the system less reliant on the human. An open-source implementation of our framework is available\footnote{\url{github.com/ori-goals/lfd-min-human-effort}}.

\subsection{A General Approach for Controller Selection}
\begin{figure}[t!]
	\centering
	\includegraphics[trim={0cm 0cm 0cm 0cm},clip,width=0.45\textwidth]{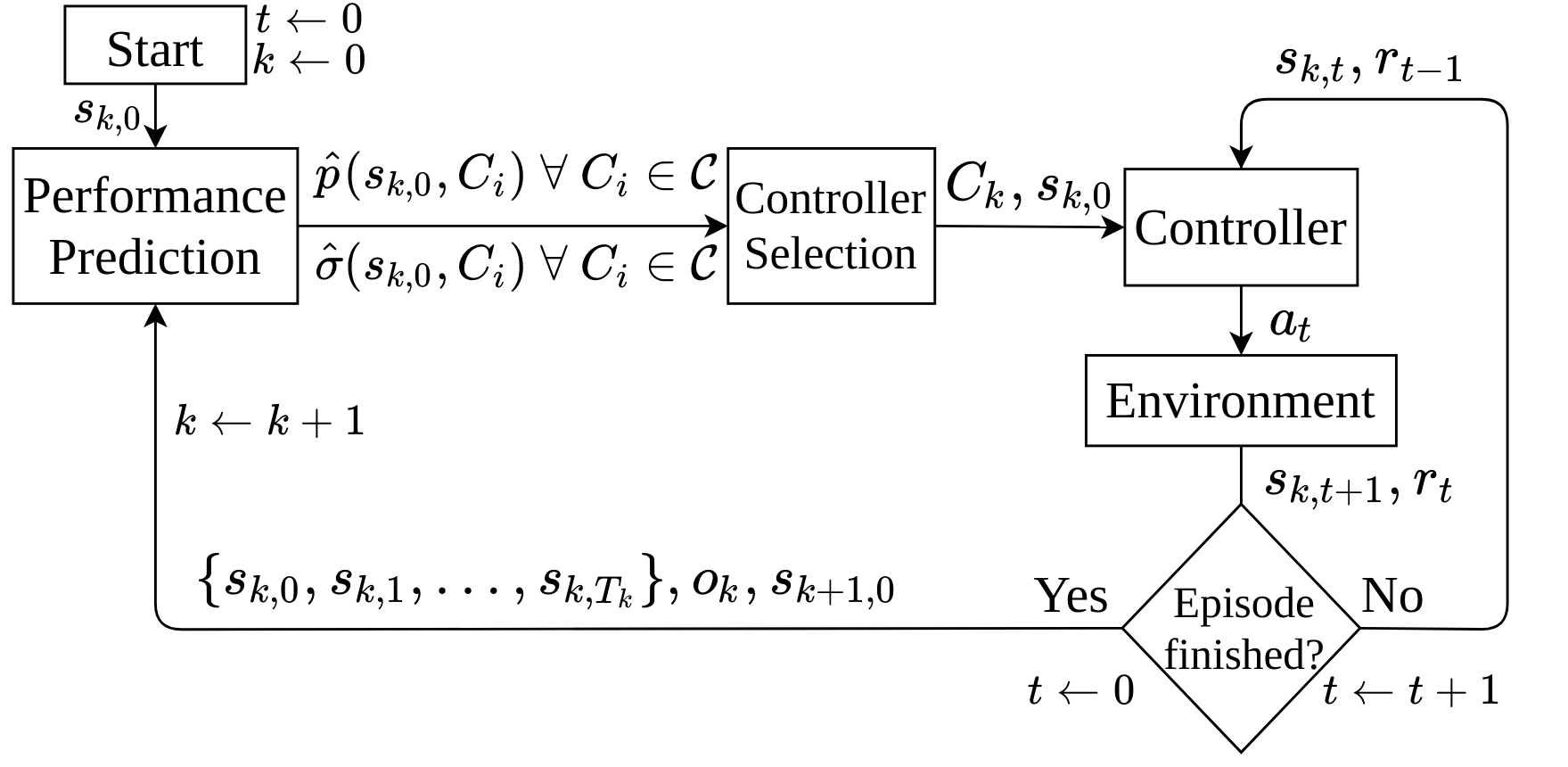} 
	\caption{High-level diagram of framework.\label{fig:framework}}
\end{figure}

Our approach to controller selection is illustrated in Figure~\ref{fig:framework}. We assume the existence of a performance prediction function, that for each controller $C_i \in \mathcal{C}$, and an initial state $s_{k,0}$ returns the probability of success for that controller for the episode, $\hat{p}(s_{k,0}, C_i)$, and the standard deviation of that probability estimate, $\hat{\sigma}(s_{k,0}, C_i)$. One possibility for this function, based on CCBPs, will be described in Section~\ref{sec:ccbp_performance}.

These estimates are then used to perform controller selection. The controller selection problem is considered as a contextual MAB, in which we seek to minimise the total human cost. To choose a controller for episode $k$, we select

\begin{equation}
C_k = \argmin_{C_i \in \mathcal{C}} \hat{c}_h(s_{k,0}, C_i),
\label{eqn:ucb_shared_autonomy}
\end{equation}

\noindent where

\scriptsize
\begin{equation}
\hat{c}_h(s_{k,0}, C_i) = 
\begin{dcases*} 
(1-\hat{p}(s_{k,0}, C_i) - \alpha \hat{\sigma}(s_{k,0}, C_i) )c_f + c_d& if  $C_i = C_h$\\ 
(1-\hat{p}(s_{k,0}, C_i) - \alpha \hat{\sigma}(s_{k,0}, C_i) )c_f& otherwise.
\end{dcases*} 
\end{equation}
\normalsize

Intuitively, $\hat{c}_h(s_{k,0}, C_i)$ is an optimistic lower bound on the cost for using controller $C_i$. This is analogous to UCB algorithms~\cite{li2010contextual}, but here we minimise cost, rather than maximise reward, and include the demonstration cost. 

The chosen controller executes actions in the environment, until the episode ends. During each step, the environment produces a reward, $r_t$, which may be used to train the controllers. At the end of the episode, the states encountered during the episode, and the outcome of success or failure, $o_k$, are fed back to improve the performance prediction function. The environment chooses a new starting state, $s_{k+1,0}$, for the next episode. This framework is agnostic to the method for performance prediction, or the controllers available. In the following subsections, we describe the use of the CCBP for performance prediction, and DDPG with demonstrations as an approach to learning one of the controllers.

\subsection{Performance Prediction with the CCBP}
\label{sec:ccbp_performance}
Under the assumption that the probability of succeeding at an episode for a given controller varies smoothly throughout the state space, we use a CCBP to estimate this probability, and its uncertainty. During episode $k$ which is under the control of $C_k$, we observe a series of states from the continuous state space \{$s_{k,0}, s_{k,1}, \ldots, s_{k,T_k}$\}. At the end of the episode we observe outcome $o_k^{C_k}$, where we introduce the superscript to indicate the controller for that episode. We consider the outcomes of episodes as outcomes of correlated Bernoulli trials. In episode $k$, we observe the same outcome for each state, ${O_k= \{B_{s_{k,0}}=o_k^{C_k}, B_{s_{k,1}}=o_k^{C_k},\ldots, B_{s_{k,T_k}}=o_k^{C_k}}$\}. After $n+1$ episodes, the entire set of observed outcomes is defined as ${O = \bigcup _{k=0}^{n} O_k}$. For a given controller $C_i$, we define the set of outcomes associated with that controller as ${O^{C_i} = \{B_{s_{k,j}} =o_k^{C_k} \in O\ |\ C_k = C_i\}}$

Given a previously unvisited state, $s$, we can calculate a beta distribution over $\Pr(B_{s} = 1\ |\ O^{C_i})$ using a CCBP as defined in Equation~\ref{eqn:ccbp}, for each $C_i \in \mathcal{C}$. In the CCBP, we initialise $\alpha(s) = \alpha_{0}^{C_i}$, and $\beta(s) = \beta_{0}^{C_i}$. This encodes a prior assumption about the probability of success of $C_i$ in the absence of information from correlated outcomes. The contributions from the outcomes in $O^{C_i}$ are then incorporated per Equation~\ref{eqn:ccbp} to produce the new beta distribution. For the success probability estimate, we take the expected value of this beta distribution: ${\hat{p}(s, C_i) = \mathbb{E}[\Pr(B_{s} = 1\ |\ O^{C_i})]}$. The standard deviation of the probability estimate is also calculated from the beta distribution: ${\hat{\sigma}(s, C_i)^2 = var[\Pr(B_{s} = 1\ |\ O^{C_i})]}$.

For the kernel function we chose the Gaussian kernel

\begin{equation}
K(s, s') = \large{ e^{ - \frac{||s - s'||^2}{l}}},
\end{equation}

\noindent where $l$ is the length scale hyperparameter which determines the scale over which the correlation between outcomes decays with distance in the state space. The Gaussian kernel was chosen as it is a common choice for modelling smooth data.

In our framework, we also allow for a controller to be learnt online from demonstrations and RL. In this case, the performance of the controller is constantly changing, and so the performance prediction should not be influenced by out of date data. We follow the approach taken for GP value functions~\cite{jakab2011improving} and non-stationary MABs~\cite{garivier2008on}, and only consider recent data in the CCBP. This is based on the assumption that the policy changes gradually, such that recent outcomes approximate the performance of the current controller. Specifically, if $C_l$ is a controller which is learnt online, then the outcomes we consider are those from the set ${{O^{C_l} = \{B_{s_{k,j}} = o_k^{C_k} \in O\ |\ C_k = C_l,\ k > n-m\}}}$, where $n$ is the current episode number, and $m$ is a constant. That is, $O^{C_l}$ is the set of outcomes from episodes controlled by $C_l$ in a sliding window of the most recent $m$ episodes. If there are many recent episodes from other controllers, $O^{C_l}$ is populated with fewer outcomes. This increases $\hat{\sigma}(s, C_l)$, indicating our reduced certainty in our probability estimates for $C_l$ after many recent demonstrations.

To simplify our experiments we assume that the human teleoperation controller is always successful, that is: ${\hat{p}(s, C_h) = 1}$ and ${\hat{\sigma}(s, C_h) = 0}$ for all $s \in S$. However, this assumption is not necessary to apply our framework, and the human performance could also be predicted with a CCBP.

\subsection{DDPG with Demonstrations}
\label{sec:ddpgwd}
To provide an autonomous controller which is learnt online from demonstrations and RL, we adapt the approach from~\cite{nair2018overcoming} to incorporate demonstrations into DDPG using behaviour cloning and a \textit{Q-filter}. As DDPG is an off-policy algorithm, we add experience from all $C_i \in \mathcal{C}$ into the replay buffer $R$. Experience from successful episodes by controllers $C_i \neq C_l$ are also added into a demonstration replay buffer $R_D$.

An update is performed after each time step of any episode. During each update, we draw $N$ experience tuples from $R$, and $N_D$ tuples from $R_D$. The following Behaviour Cloning (BC) loss, $L_{BC}$, is applied only to the tuples from $R_D$:

\begin{equation}
L_{BC} = f\sum_{i=1}^{N_D}||\pi(s_i|\theta^\pi) - a_i||^2,
\end{equation}
\begin{equation}
f = \mathbb{1}_{Q(s_i, a_i) > Q(s_i, \pi(s_i)) - \epsilon|Q(s_i, \pi(s_i))|},
\label{eqn:filter}
\end{equation}

\noindent where $\epsilon \geq 0$, $a_i$ is the demonstrator action, and $\mathbb{1}_A$ is 1 if $A$ is true and 0 otherwise. The Q-filter in Equation \ref{eqn:filter} results in the behaviour cloning loss not being applied when the critic determines that the actor action is significantly better than the demonstrator action. This prevents the actor being tied to the demonstrations if it discovers better actions. This Q-filter is modified from~\cite{nair2018overcoming} by including the $\epsilon$ term, which when $\epsilon > 0$ reduces the experiences filtered out from the BC loss.

The gradient applied to $\theta^\pi$ is:
\begin{equation}
\lambda_{1} \nabla_{\theta^\pi}J - \lambda_{2}\nabla_{\theta^\pi} L_{BC},
\end{equation}

\noindent where $\lambda_1$ and $\lambda_2$ weight the policy gradient and behaviour cloning contributions to the update.

\section{Experiments}
We evaluated our approach in three domains. We assumed the availability of three controllers: ${\mathcal{C} = \{C_h, C_b, C_l\}}$. $C_h$ was a human teleoperating the robot with a gamepad. $C_b$ was a pre-programmed baseline controller capable of succeeding at some of the episodes. $C_l$ was a control policy learnt online using the method outlined in Section~\ref{sec:ddpgwd}. In all domains, the maximum length of an episode was 50 time steps, and the sliding window was $m=50$ episodes. We set $\alpha_0^{C_i}$, $\beta_0^{C_i}$ to define a prior assumption that ${\hat{p}(s_{k,0}, C_l) = \hat{p}(s_{k,0}, C_b) = 0.8}$, and ${\hat{\sigma}(s_{k,0}, C_l) = \hat{\sigma}(s_{k,0}, C_b) = 0.35}$, in the absence of any observed outcomes ($\alpha_0^{C_i} = 0.245$, $\beta_0^{C_i} = 0.0612$).  To break ties in Equation~\ref{eqn:ucb_shared_autonomy}, we prioritised $C_b > C_l > C_h$.

\subsection{Domains}
\subsubsection{Block Pushing}
A 6-DOF arm was simulated in Gazebo~\cite{koenig2004design} to push a block from a random initial configuration along a table into a goal region (see Figure~\ref{fig:domains}). The state space is 32 dimensional, consisting of 30 depth measurements from a stationary laser, and 2 measurements specifying the current location of the end effector. The action space is 2 dimensional, specifying a horizontal change in position of the end effector. In each new episode, the arm position is reset, and a block is spawned onto a $0.3$m$\times0.45$m table. The block has a 0.04m$\times$0.04m cross section and can be one of three lengths: 0.12, 0.2, or 0.3m. The initial position and orientation of the block is varied by up to $\pm10$cm relative to the centre of the table, and $\pm20^\circ$. An episode is a failure if the block falls off the table, or the time limit is exceeded. An episode is a success if the entirety of the block is pushed past the green line (Figure~\ref{fig:domains}). The baseline controller, $C_b$, was a partially trained policy using the method in Section~\ref{sec:ddpgwd} and a sample of the recorded human demonstrations. Training of $C_b$ was terminated when the policy was capable of $\approx$70\% of the tasks. The reward $r_t$ used to train the learnt policy consisted of a small dense reward for moving the block forwards, and a large reward for reaching the goal region. 

\subsubsection{Simulated Navigation}
A Scitos G5\footnote{\url{metralabs.com/en/mobile-robot-scitos-g5/}} robot was simulated in the Morse simulator~\cite{morse_simpar_2012}. The task for the robot was to navigate from a starting location, through a gap, to a goal region. 
The starting location was randomly selected within a region. The start and goal regions are shown in Figure~\ref{fig:domains}. The starting orientation was randomly varied $\pm10^\circ$. The width of the gap was varied according to a normal distribution with a mean of 0.83m (the standard wheelchair accessible door width in the United Kingdom), a standard deviation of 0.15m, and a lower bound of 0.68m. This lower bound is the minimum width at which the 0.62m robot can be reliably teleoperated through the gap. The position of the other walls was randomly varied by $\pm0.25$m.
The state space was a 40 dimensional laser-depth measurement, and the action space was 2 dimensional, specifying a distance to move and a change in orientation.
The baseline controller, $C_b$, was a controller from the ROS navigation stack with the parameters fine-tuned for the Scitos G5 from the STRANDS project\footnote{\url{github.com/strands-project/strands\_movebase}}.
The reward $r_t$ used to train the learnt policy consisted of a large reward for reaching the goal region, and 0 otherwise.  

\subsubsection{Real-World Navigation}
The real-world navigation experiments used a Turtlebot 2\footnote{\url{clearpathrobotics.com/turtlebot-2-open-source-robot/}}. The task was to navigate from a starting location and through a partially closed door. Possible starting locations are indicated by the region marked on the floor in Figure~\ref{fig:domains}. To configure a new episode, the ROS navigation stack was used to navigate the robot to a random position within this region, and a random orientation varied up to~$\pm15^\circ$. The door was set randomly to one of 8 possible positions indicated by markings on the floor. Three possible door positions are shown in Figure~\ref{fig:domains}. The state space is 33 dimensional, comprising of 30 depth measurements given by the Kinect sensor on the Turtlebot, and 3 measurements specifying the estimated position and orientation of the robot given by a particle filter. The particle filter estimate is required because the Kinect has a narrow field of view, meaning that the depth measurements alone are insufficient to characterise the state. The action space was 2 dimensional, consisting of linear and angular velocity. The reward was a sparse reward for passing through the door, and 0 otherwise.

\subsection{Length Scale Hyperparameter Estimation}
\label{sec:hyperparam}
The hyperparameter $l$ was estimated from data in all domains. We initialised a policy $\pi$ capable of completing approximately half of the tasks. The policy attempted 50 random tasks to populate $O$. The policy then executed another 50 episodes from new random initial states, which were not added to $O$. The likelihood of the outcomes of the second set of tasks was calculated using the CCBP for a range of values for $l$. The maximum likelihood estimate was then used for the CCBP kernel for both $C_l$ and $C_b$. This value was $l=4.1$ in the simulated navigation domain, $l=2.1$ in the real-world navigation domain, and $l=0.72$ in the block pushing domain. We leave estimating this parameter online to future work.\\

\subsection{Training Details}
To train $C_l$, we used Adam~\cite{kingma2014adam} with learning rate of $10^{-3}$ for $Q$, and $10^{-4}$ for $\pi$. The discount factor $\gamma$ was 0.99. We used ${N = 128}$, ${N_D = 64}$, $\lambda_1 = 1$, $\lambda_2 = 10$. The value for $\lambda_2$ was chosen for best empirical performance after 100 demonstrations and 500 RL episodes out of $\{1, 10, 100\}$ in block pushing. The function approximators for $\pi$ and $Q$ are fully-connected neural networks with two hidden layers of 64 and 32 weights respectively, and ReLU activation. The output activation for $\pi$ is tanh, and the value is rescaled into the action range. The exploration noise $\mathcal{N}$ is an Ornstein-Uhlenbeck process with $\sigma = 0.2$, $\theta=0.15$. The noise was decayed by $0.998^{n_{C_l}}$, where $n_{C_l}$ is the number of episodes completed by ${C_l}$. The Q-filter tolerance was~${\epsilon = 0.02}$ unless stated otherwise.

\subsection{Simulated Human Cost Evaluation}
We evaluated the cumulative human cost of several methods in simulation. The costs were $c_d = 1$ and $c_f =5$ for both simulated domains. $n$ random initial states were generated for each simulated domain. In box pushing, $n = 1200$ and in simulated navigation, $n=500$. A human demonstration for each of the $n$ possible initial states was recorded. Each run of the experiment consisted of randomly initialising the networks for $C_l$, and then performing a number of episodes (400 or 1200 in simulated navigation or box pushing respectively). In each run, the initial state for each episode was chosen in a random order from the set of possible initial states. In each run, the same recorded human demonstration was used for each initial state. We compared several methods, where $\alpha$ indicates the value used in the MAB: \textit{contextual MAB ($\alpha$)}  is our approach with ${\mathcal{C} = \{C_h, C_l\}}$, \textit{contextual MAB with baseline ($\alpha$)} is our approach with ${\mathcal{C} = \{C_h, C_b, C_l\}}$, \textit{baseline only} always uses $C_b$ and \textit{RL only} always uses $C_l$. The method \textit{human then learner} ($n_h$), first executes $n_h$ human demonstrations with $C_h$ before switching to $C_l$. For \textit{Boltzmann ($\Delta \tau$)}, ${\mathcal{C} = \{C_h, C_b, C_l\}}$,  and we adapted the method in~\cite{fernandez2006probabilistic}. The controller is chosen according to:
\begin{equation}
Pr(C_i)  = \frac{e^{\tau(c_f - \bar{c}_h(C_i))}}{\sum_{C_j \in \mathcal{ C}} e^{\tau(c_f - \bar{c}_h(C_j))}}
\end{equation}

\noindent where $\bar{c}_h(C_i)$ is the average human cost incurred from the episodes in $O^{C_i}$. Temperature parameter $\tau$ is initially 0, and incremented by $\Delta \tau$ every episode.

\begin{figure}
	
	\centering
	\begin{subfigure}[b]{0.46\textwidth}
		\includegraphics[width=1\linewidth]{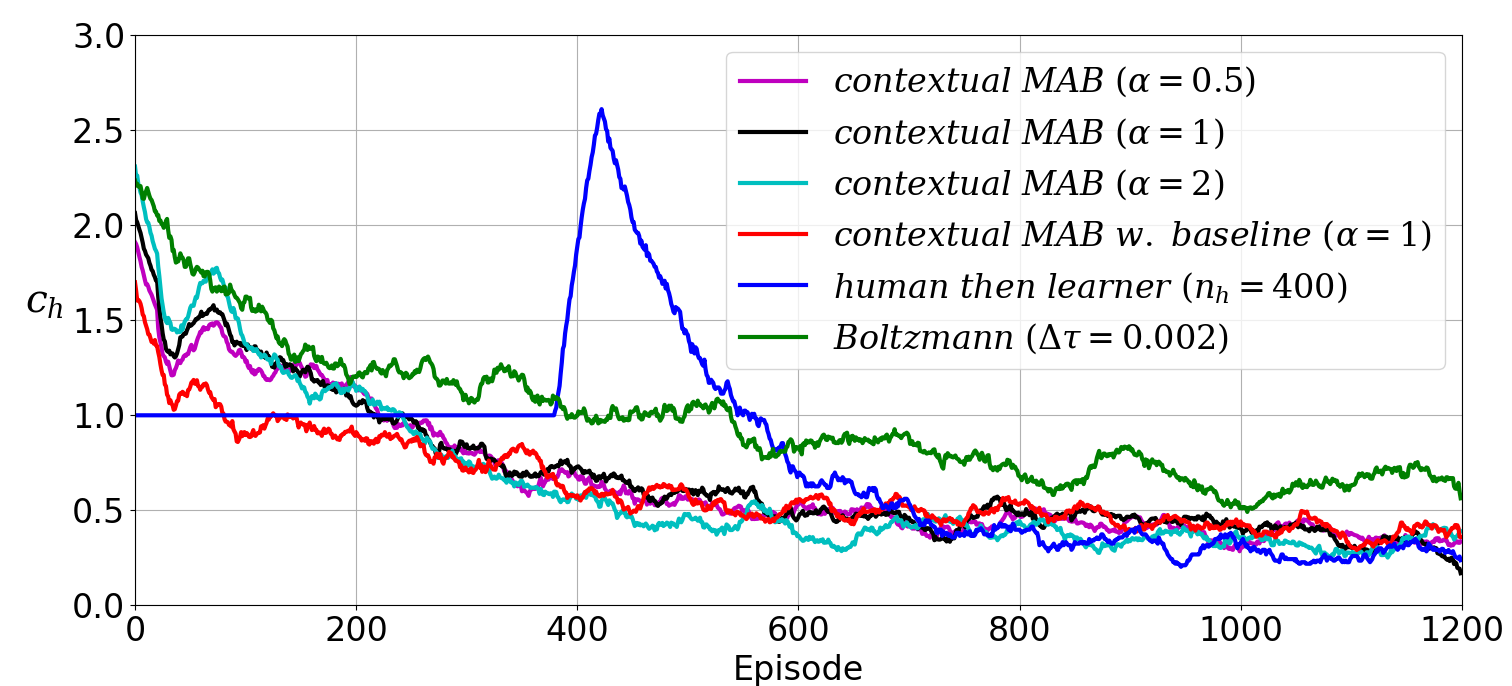}
		\caption{Block pushing}
		\label{fig:Ng1} 
	\end{subfigure}
	
	\begin{subfigure}[b]{0.46\textwidth}
		\includegraphics[width=1\linewidth]{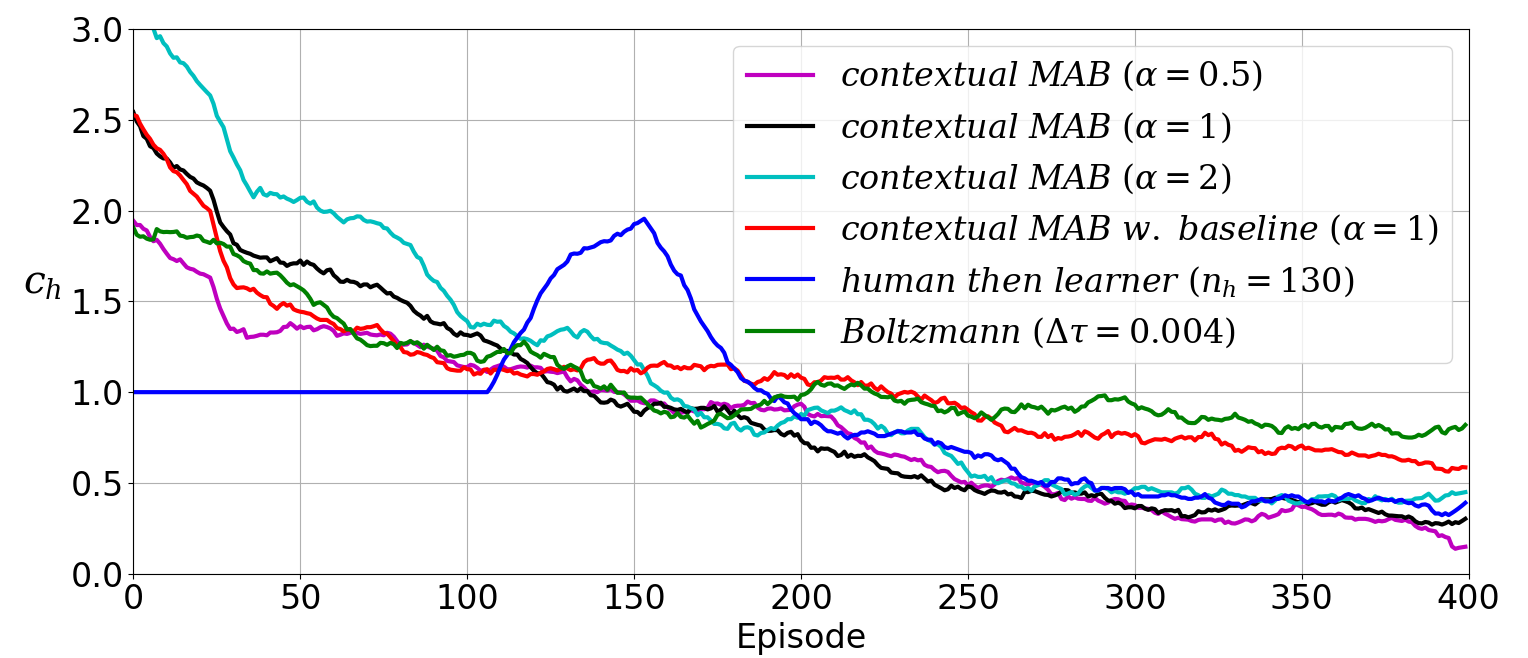}
		\caption{Simulated Navigation}
		\label{fig:Ng2}		
	\end{subfigure}
	
	\caption{Mean cost from 15 runs of each method. For clarity, $c_h$ is plotted using a sliding window average over 40 episodes.\label{fig:human_cost}}
	\vspace{-1mm}
\end{figure}

\textit{Results:} The average human cost over 15 runs of each method is plotted in Figure~\ref{fig:human_cost}. For clarity, the cost is plotted using a sliding window average over 40 episodes. The average cumulative cost and episodes performed by each controller is displayed in Tables~\ref{tab:table1} and~\ref{tab:table2}. In block pushing, \textit{contextual MAB} outperforms \textit{human then learner} in total cost for all values of $\alpha$ and $n_h$. Smaller $\alpha$ values corresponded with asking for more demonstrations. The method \textit{contextual MAB with baseline} further reduces the cost by using the baseline controller when the learnt policy is poor. Despite the fact that \textit{baseline only} does not perform well, this is possible as \textit{contextual MAB with baseline} tends to use the baseline for episodes it will likely succeed at. The \textit{Boltzmann} method performs comparatively poorly for each of the values of $\Delta \tau$.

\begin{table}[t!]
	\begin{center}
		\caption{Cumulative cost over 15 runs of each method in the box pushing domain in the format: mean (std dev.)}
		\label{tab:table1}
		\resizebox{0.96\columnwidth}{!}{%
			\begin{tabular}{|l|c|c|c|} 
				\hline
				\textbf{Method} & \begin{tabular}[c]{@{}c@{}}\textbf{Total}\\\textbf{Cost}\end{tabular} & \begin{tabular}[c]{@{}c@{}}\textbf{Human}\\\textbf{Episodes}\end{tabular} & \begin{tabular}[c]{@{}c@{}}\textbf{Baseline}\\\textbf{Episodes}\end{tabular}\\
				\hline
				\textit{context. MAB ($\alpha = 0.5$)} & 790 (117) &  376 (76) & - \\
				\hline
				\textit{context. MAB ($\alpha = 1$)} & 820 (119) &  332 (73) & - \\
				\hline
				\textit{context. MAB ($\alpha = 2$)} & 758 (167) &  220 (67) & - \\
				\hline
				\textit{human then learner ($n_h = 200$)} &  1029 (90) & 200 & - \\
				\hline
				\textit{human then learner ($n_h = 300$)} & 959 (133) & 300 & - \\
				\hline
				\textit{human then learner ($n_h = 400$)} & 934 (94) & 400 & - \\
				\hline
				\textit{context. MAB w. baseline ($\alpha = 1$)} & 746 (75) & 197 (32)& 236 (35)\\
				\hline
				\textit{Boltzmann ($\Delta \tau = 0.001$)} & 1266 (74)  & 392 (15) & 327 (25) \\
				\hline
				\textit{Boltzmann ($\Delta \tau = 0.002$)} & 1178 (94)  & 364 (30) & 277(35)\\
				\hline
				\textit{Boltzmann ($\Delta \tau = 0.004$)} & 1204 (188) & 500 (221) & 244 (67)\\
				\hline
				\textit{baseline only} & 1613 (18) & - & 1200 \\
				\hline
				\textit{RL only} & 4947 (39) & - & - \\
				\hline
			\end{tabular}
		}
	\end{center}
	\vspace{-2mm}
\end{table}

\begin{table}[h!]
	
	\small
	\begin{center}
		\caption{Cumulative cost over in the simulated navigation domain in the format: mean (std dev.)}
		\label{tab:table2}
		\resizebox{0.96\columnwidth}{!}{%
			\begin{tabular}{|l|c|c|c|} 
				\hline
				\textbf{Method} & \begin{tabular}[c]{@{}c@{}}\textbf{Total}\\\textbf{Cost}\end{tabular} & \begin{tabular}[c]{@{}c@{}}\textbf{Human}\\\textbf{Episodes}\end{tabular} & \begin{tabular}[c]{@{}c@{}}\textbf{Baseline}\\\textbf{Episodes}\end{tabular}\\
				\hline
				\textit{context. MAB ($\alpha = 0.5$)} & 358 (51) & 144 (33)  & - \\
				\hline
				\textit{context. MAB ($\alpha = 1$)} & 366 (48) & 110 (30)  & - \\
				\hline
				\textit{context. MAB ($\alpha = 2$)} & 438 (53) & 67 (16)  & - \\
				\hline
				\textit{human then learner ($n_h = 70$)} &  517 (147) &  70 & - \\
				\hline
				\textit{human then learner ($n_h = 100$)} &  409 (75) & 100 & - \\
				\hline
				\textit{human then learner ($n_h = 130$)} &  349 (71) & 130 & - \\
				\hline
				\textit{context. MAB w. baseline ($\alpha = 1$)} & 433 (72) & 57 (19) & 118 (47)\\
				\hline
				\textit{Boltzmann ($\Delta \tau = 0.001$)} & 499 (46)  & 137 (9) & 133 (11) \\
				\hline
				\textit{Boltzmann ($\Delta \tau = 0.002$)} & 505 (50)  & 155 (17) & 147 (24) \\
				\hline
				\textit{Boltzmann ($\Delta \tau = 0.004$)} & 473 (56)  &  153 (27) & 162 (26) \\
				\hline
				\textit{baseline only} & 443 (14)& - & 400 \\
				\hline
				\textit{RL only} & 1811 (46) & - & - \\
				\hline
			\end{tabular}
		}
	\end{center}
	\vspace{-3mm}
\end{table}

In the simulated navigation domain, \textit{contextual MAB} performs better with a smaller value for $\alpha$. This is likely because this simpler task can be learnt from demonstrations alone and no RL experience. Therefore, conservatively relying on more human demonstrations enables a good policy to be learnt quickly and results in less cost. This also explains the strong performance of \textit{human then learner} ($n_h = 130$). This suggests that our approach is better suited to more complex tasks which take longer to learn, and require RL experience to train a policy to complete the task. The method \textit{contextual MAB with baseline} has less cost near the start of the runs, but performs worse overall. This is likely because the agent is slower to learn a good policy using the dissimilar demonstrations from the baseline controller and human in the simulated navigation domain. In the box pushing domain, the baseline and human give similar demonstrations because the baseline is a policy learnt from human demonstrations. Therefore this is not an issue in the box pushing domain.

\subsection{Effect of Varying Costs}
Some experiments were repeated in the block pushing domain to analyse the effect of the cost values. The demonstration cost, $c_d = 1$ for all experiments. We used $c_f \in \{3, 5, 7\}$ for \textit{contextual MAB} and \textit{human then learner} with a comparable number of demonstrations.

\textit{Results}:
Our approach incurs less cost than \textit{human then learner} for all values of $c_f$ (Table~\ref{tab:cost_anal}). Increasing the value of $c_f$ increases the number of human demonstrations used and vice versa. This is because if the cost for failure is higher the estimated probability of success must be higher for the algorithm to choose the autonomous controller.

\begin{table}[h!]
	
	\small
	\begin{center}
		\caption{Cost over 15 runs of 1200 episodes for block pushing with different $c_f$ values in format: mean (std dev.)}
		\label{tab:cost_anal}
		\resizebox{0.9\columnwidth}{!}{%
			\begin{tabular}{|l|c|c|c|} 
				\hline
				\textbf{Method} & $c_f$ & \begin{tabular}[c]{@{}c@{}}\textbf{Total}\\\textbf{Cost}\end{tabular} & \begin{tabular}[c]{@{}c@{}}\textbf{Human}\\\textbf{Episodes}\end{tabular}\\
				\hline
				\textit{context. MAB ($\alpha = 1$)} & 3 & 629 (82) & 281 (52) \\
				\hline
				\textit{human then learner ($n_h = 300$)} & 3  & 715 (57) & 300 \\
				\hline
				\textit{context. MAB ($\alpha = 1$)} & 5 & 820 (119) & 332 (73) \\
				\hline
				\textit{human then learner ($n_h = 300$)} & 5 & 959 (133) & 300 \\
				\hline
				\textit{context. MAB ($\alpha = 1$)} & 7 & 1067 (121)  & 421 (74) \\
				\hline
				\textit{human then learner ($n_h = 400$)} &7  & 1154 (122)  & 400 \\
				\hline
			\end{tabular}
		}
	\end{center}
	\vspace{-5mm}
\end{table}

\subsection{Limited Demonstrations Evaluation}
\begin{figure}[h!]
	\centering
	\includegraphics[trim={0cm 0cm 0cm 0cm},clip,width=0.45\textwidth]{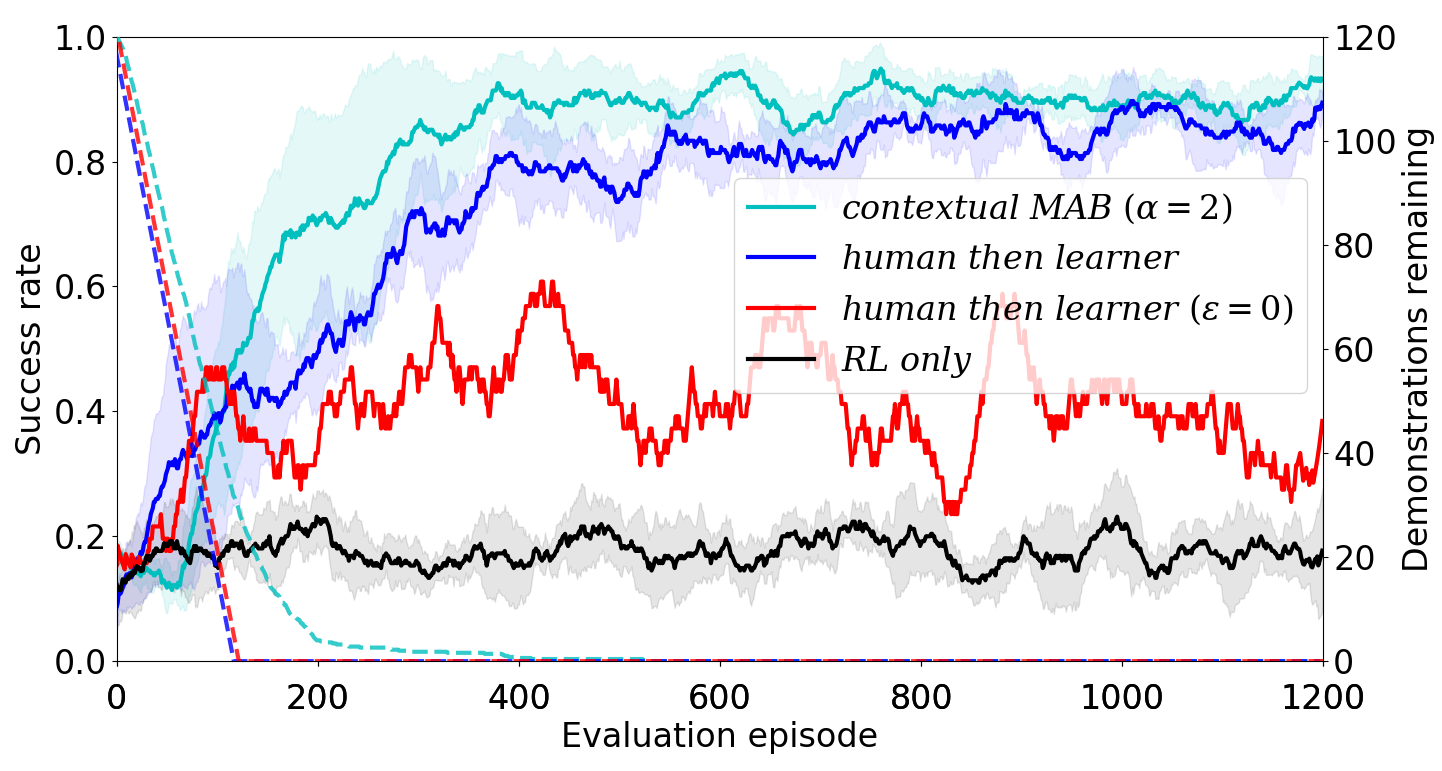} 
	\caption{Success rate on evaluation episodes for policies trained with a limited budget of 120 human demonstrations. Results are averaged over 5 runs for each method. Solid lines indicate average success rate over a sliding window of 40 episodes. Shaded regions illustrate the standard deviation over the 5 runs. Dashed lines indicate the demonstrations remaining.\label{fig:prelim}}
	\vspace{-3mm}
\end{figure}

We evaluated the quality of the learnt policy when the number of human demonstrations is kept to a strict limit. After the demonstration budget was used up, $C_l$ was used from then onwards. To evaluate policy performance, after each episode we performed an additional evaluation episode which always used $C_l$. The initial state for each evaluation episode was sampled randomly. The success rate of $C_l$ in the evaluation episodes is plotted in Figure~\ref{fig:prelim} for block pushing, where the limit on human demonstrations was 120. We additionally compared \textit{human then learner} with $\epsilon = 0$ for the Q-filter.

\textit{Results}: 
Early in training, \textit{human then learner} outperforms \textit{contextual MAB} as it receives all demonstrations immediately. However, \textit{contextual MAB} converges to a policy with a high success rate more quickly. This indicates that \textit{contextual MAB} received more informative demonstrations by asking for demonstrations at initial states where a good policy had not yet been learnt. \textit{RL only} failed to learn a good policy in the limited number of episodes. Our results show that setting $\epsilon=0$ for the Q-filter resulted in slower learning by filtering out most demonstration experience from the behaviour cloning loss. A promising approach may be to start with a large value for $\epsilon$, and then gradually decay $\epsilon$. 

\subsection{Real-World Experiment}
In the real-world experiment we compared \textit{contextual MAB} and \textit{human then learner} over one run each of 400 episodes. In the real-world experiments, we used $\epsilon = 0.1$ and a larger neural network with hidden layers of 128 and 64 weights as this was empirically found to improve the performance of the learnt policies. All other training parameters were the same as for the simulated experiments. For \textit{contextual MAB} we used $\alpha = 1$. The costs were $c_d = 1$ and $c_f =5$. 

\begin{figure}[h!]
	\centering
	\includegraphics[trim={0cm 0cm 0cm 0cm},clip,width=0.45\textwidth]{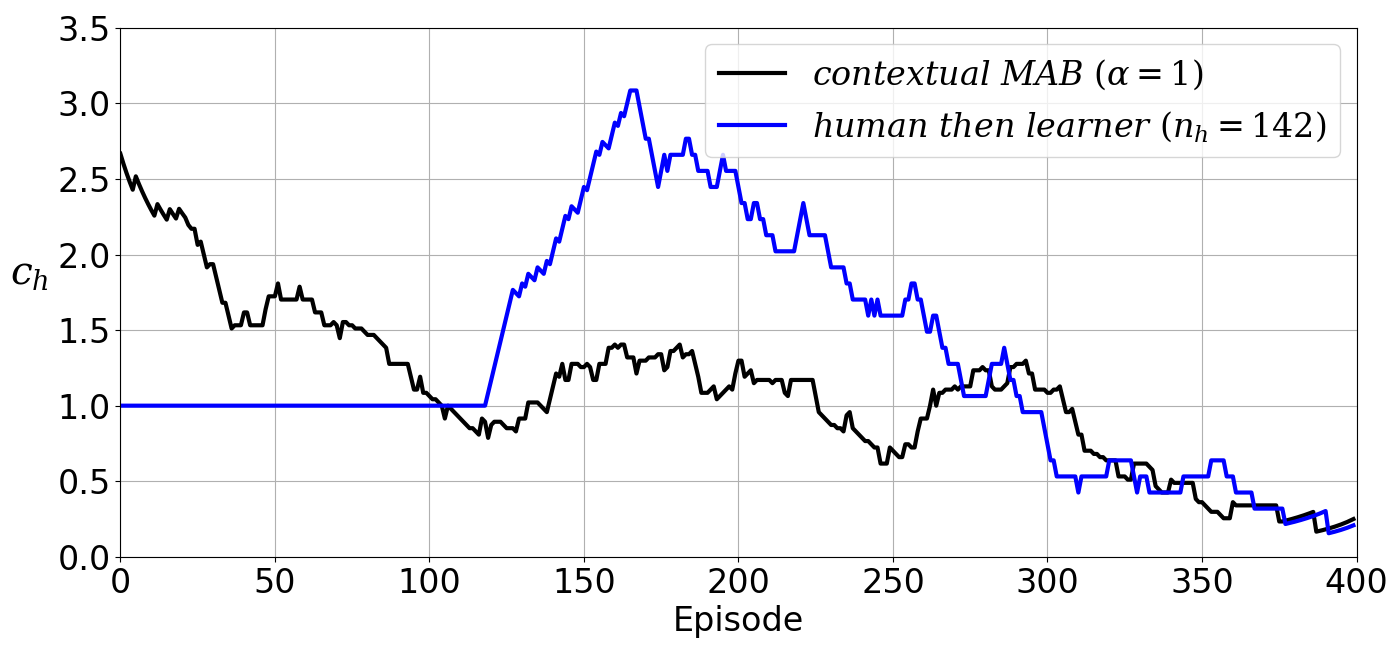} 
	\caption{Human cost in real-world navigation experiment. $c_h$ is plotted using a sliding window average over 40 episodes.\label{fig:turtlebot_cost}}
	\vspace{-3mm}
\end{figure}

\textit{Results}:
\textit{contextual MAB} used 142 human demonstrations and incurred a total cost of 442 over the 400 total episodes. With exactly the same number of demonstrations, \textit{human then learner} incurred more failures, accumulating a total cost of 517. The cost incurred throughout the experiment is plotted in Figure~\ref{fig:turtlebot_cost}. These results demonstrate that our approach can easily be applied to real-world robotics problems.

\subsection{Discussion}
The $\alpha$ value in our approach enables the user to tune whether the system conservatively asks for more demonstrations, or explores trying the other controllers. A smaller value reduces the risk of failures, but with a higher value the system more quickly gathers RL experience and more accurately predicts controller performance. Our results from our two simulated domains indicate that simple tasks that can be learnt from demonstrations alone favour a smaller value for $\alpha$ to conservatively ask for demonstrations and avoid failures. For more difficult problems which take longer to learn and require RL experience to train a good policy, a larger value may be favourable to gather more RL experience. 

Choosing the controller for each episode based on context is key to the success of our approach. The \textit{Boltzmann} method favours choosing the learnt policy when it begins to perform well overall. However, our method performs better by selectively favouring the autonomous controllers only from initial states where they are likely to succeed. As a side effect, our approach tends to ask for more informative demonstrations, resulting in better performance of the learnt policy with the same number of demonstrations. 

Our results indicate that including demonstrations from different sources increases the amount of experience required to learn a good policy. It may be useful to use our approach to learn from many different controllers (for example, a number of different teleoperators who have different techniques). However, the issue of learning effectively from dissimilar demonstrations may need to be addressed before our approach is advantageous with a large number of different controllers.

\section{Conclusion}
We have presented an approach to choosing between human teleoperation and autonomous controllers to minimise the cost of bothering the human operator. By estimating the performance of each controller throughout the state space, our system reduces the human cost by only asking for demonstrations when they are needed, and reducing autonomous failures. By learning one of the autonomous controllers online, our system becomes less reliant on the human with more experience. In future work, we will apply the framework presented in this paper to problems where the human operator may make mistakes, and so we will also need to predict human performance. Additionally, we will extend our approach to sequential decision making problems to consider \textit{sequences} of controller selections. To further reduce the human effort required, our framework could be applied using more data-efficient learning methods for the learnt controller, such as model-based RL methods.

\bibliographystyle{IEEEtran}
\bibliography{IEEEabrv,mybib}
%
%




\end{document}